\documentclass[11pt,a4paper]{article}
\title{A survey of diversity quantification in natural language processing: \\ The why, what, where and how}
\author{
{L}ouis {E}st\`eve \\
France \\
Universit\'e Paris-Saclay, CNRS, LISN\thanks{\texttt{first.last@universite-paris-saclay.fr}} \\
\And
{M}arie-{C}atherine de {M}arneffe \\
Belgium \\
FNRS - UCLouvain\thanks{\texttt{marie-catherine.demarneffe@uclouvain.be}} \\
\And
{N}urit {M}elnik \\
Israel \\
The Open University of Israel\thanks{\texttt{nuritme@openu.ac.il}}
\AND
{A}gata {S}avary \\
France \\
Universit\'e Paris-Saclay, CNRS, LISN\footnotemark[1]
\And
{O}lha {K}anishcheva \\
Germany \\
Heidelberg University\thanks{\texttt{kanichshevaolga@gmail.com}}
}
\date{}
\usepackage{times,latexsym} 
\usepackage{soul}
\usepackage{url} \usepackage[T1]{fontenc}
\usepackage{booktabs}
\usepackage{array}
\usepackage[acceptedWithA]{tacl2021v1}
\usepackage{xspace,mfirstuc,tabulary}   
\newif\iftaclinstructions \taclinstructionsfalse \iftaclinstructions
\renewcommand{\confidential}{} \renewcommand{\anonsubtext}{(No author info
supplied here, for consistency with TACL-submission anonymization requirements)}
\newcommand{\instr} \fi \iftaclpubformat  
   \else

 \fi \usepackage{graphicx}
\usepackage{tikz}
\usetikzlibrary{shapes,shapes.geometric,arrows,positioning,calc}
\usepackage{longtable}
\usepackage{natbib} 
\usepackage{amsmath} 
\usepackage{microtype}

 \newcommand{\one}{1} \newcommand{\enablecomments}{1}
\newcommand{\mcdm}[1]{\ifx\enablecomments\one{}\textbf{\color{blue}[mcdm:{}#1]}{}\fi}
\newcommand{\louis}[1]{\ifx\enablecomments\one{}\textbf{\color{brown}[louis:{}#1]}{}\fi}
\newcommand{\giedre}[1]{\ifx\enablecomments\one{}\textbf{\color{yellow}[giedre:{}#1]}{}\fi}
\newcommand{\irina}[1]{\ifx\enablecomments\one{}\textbf{\color{red}[irina:{}#1]}{}\fi}
\newcommand{\agata}[1]{\ifx\enablecomments\one{}\textbf{\color{cyan}[agata:{}#1]}{}\fi}
\newcommand{\nurit}[1]{\ifx\enablecomments\one{}\textbf{\color{purple}[nurit:{}#1]}{}\fi}
\newcommand{\olha}[1]{\ifx\enablecomments\one{}\textbf{\color{orange}[olha:{}#1]}{}\fi}
\newcommand{\quotecite}[1]{\citeauthor{#1}'s (\citeyear{#1})}
\usepackage{enumitem} \setlist{nolistsep} 

\usepackage{multirow}
\usepackage{longtable}
\usepackage{supertabular}
\usepackage{comment}
\usepackage{array}
\usepackage{enumitem}

\newcommand{\final}[1]{#1} %

\begin{document}

\maketitle
\vspace{5cm}
\begin{abstract}

The concept of diversity received increasing attention in natural language processing (NLP) in recent years. It became an advocated property of datasets and systems, with many measures used to quantify it, but also many cross-paper inconsistencies. 
To address this fragmentation, we take inspiration from other scientific fields where diversity has been more thoroughly conceptualized. 
\final{We build upon \cite{sarkar_diversity_2010} and  \cite{stirling_general_2007}, two unified cross-domain frameworks inspired from ecology, economy and other domains.
The first one advocates explicit justification of diversity endorsement. The latter} distinguishes three dimensions of diversity: variety, balance, and disparity. Surveying over 300 recent diversity-related papers from ACL Anthology, we propose an NLP-specific framework with 4 perspectives: \emph{why} diversity is important, \emph{what} diversity is measured on, \emph{where} it is measured, and \emph{how}. 
\final{We find that in NLP richness (the simplest variety) %
and distance averaging (the simplest disparity) are the most prevalent measure families, and that pure balance measures are nearly absent. 
 Moreover, half of the measures used in NLP do not conform to Stirling's framework. } 
 Our analysis increases comparability of approaches to diversity in NLP, allowing us to formulate recommendations for the field. 
 \final{We recommend that the field systematically use a unified terminology, and that it engage in debates about the desirable properties of diversity measures as well as about the normative assumptions behind their use.}
\end{abstract}

\section{Introduction} \label{section:introduction}

\begin{figure}[ht!]
\centering
\resizebox{\linewidth}{!}{\input{images/exp-012_final_flat.pgf}}
\resizebox{\linewidth}{!}{\input{images/exp-012_final_ratio.pgf}}
\caption{ACL Anthology papers from 1990 to
2024 with ``diversity'' or ``diverse'' in their title
or abstract.
\label{fig:diversity-papers}}
\end{figure}

The notion of diversity has been gaining increasing attention in natural
language processing (NLP) over the past few years. The top panel of
Figure~\ref{fig:diversity-papers} depicts the number of papers in the ACL
Anthology from 1990 to 2024 which contain ``diversity'' or ``diverse'' in their
title or abstract.\footnote{Prior to 1990, the only one such item is an erratum
entitled \emph{Diverse corrigenda and addenda to the pre-prints}.} There are
4,357
such papers in total,
472
of them contain one of these
terms in their title (in orange), and
3,885
in their abstract but not in their title (in blue).
This increase is not only
related to the rapid growth of NLP in general.
The bottom panel of Figure~\ref{fig:diversity-papers} shows the
same information, but as a year-wise share.
In particular, in 2024, 13\% of new
papers contained the term ``diverse'' or
``diversity'' in their title or abstract, almost 5 points
above 2023.

To date there have been very few attempts to take a step back and understand the conceptualization of diversity in NLP and the motivations behind its endorsement. When such attempts were made, they
were limited to particular areas
\cite{%
    tevet-berant-2021-evaluating,%
    yang-etal-2025-measuring,%
    zhang-etal-2025-evaluating-evaluation%
}
and diversity aspects
\cite{%
    lion-bouton-etal-2022-evaluating,%
    ploeger-etal-2024-typological%
}.
Overall, the field currently suffers from the lack of a shared vocabulary and framework, which leads to conceptual fragmentation concerning diversity.
This suggests that NLP
belongs to the ``fields [\ldots] where diversity is prominent in discussion,
but remains undefined or analytically neglected'' \cite[p.~707]{stirling_general_2007}.

The objective of this survey is to pave the way toward addressing these shortcomings by taking inspiration from studies outside of NLP where diversity has been systematically analyzed
\citep{sarkar_diversity_2010, stirling_general_2007}.
These studies revealed
cross-domain similarities and proposed unifying conceptualizations. We examine to which extent these proposals can be applied to how the concept of diversity is used, and especially \emph{quantified}, in NLP.

Our contributions are twofold. First, we propose \textbf{an NLP-specific framework} for conceptualizing diversity quantification, which aligns with the aforementioned cross-domain analyses. It consists of four perspectives: the
motivations behind the quest for diversity (\emph{why}), the objects on which
diversity is quantified (\emph{what}), the NLP areas and pipeline stages where diversity
measures are applied (\emph{where}) and the types of diversity measures
themselves (\emph{how}). This framework, applied to over 300 diversity-related NLP papers, increases their comparability and makes global tendencies emerge. 
\final{We find that only roughly 60\% of the NLP papers surveyed  actually quantify diversity. These papers use approximately 80 measures, bearing over 100 different names. Thus, NLP currently is in a ``proliferation of measures'' stage, paralleling what has been seen in the evolution of ecology. Roughly half of these measures, which we group into 7 families, conform to the diversity dimensions advocated by \citeauthor{stirling_general_2007}. Among these, richness (counting categories) and averaging distances
are the most frequent. The other half of measures, non-conforming to \citeauthor{stirling_general_2007}'s dimensions, is grouped into 6 families, of which the type-token ratio family is the most frequent.} %

Second, we put forward \textbf{recommendations} for further conceptualization of diversity in the field. \final{We suggest that papers advocating diversity should explicitly %
address the \textit{why}, \textit{what}, \textit{where} and \textit{how} perspectives. }
\final{The community should further debate about desirable properties of diversity measures and about
the normative assumptions behind their use.
We expect these recommendations to lay the groundwork for better comparability and for educated choices of measures that are proposed in the literature, building shared and stable terminology and formalisation of diversity quantification.}

\section{Cross-domain perspective on diversity} %
\label{sec:other-fields}

\final{Diversity is a recurring concept in many disciplines, including biology, ecology, linguistics, cultural studies, economics, and policy studies. How these fields conceptualize and measure diversity may offer useful perspectives for NLP. Two works have analyzed diversity systematically across fields. \citet{sarkar_diversity_2010} asks why diversity is valued and uncovers the normative assumptions underlying its endorsement (Sec.~\ref{sec:philo}). \citet{stirling_general_2007} proposes a unified conceptualization characterizing diversity along three dimensions (Sec.~\ref{sec:stirling}). Finally, ecology, with its long history of diversity measurement, offers a detailed case study of how diversity measurement evolves (Sec.~\ref{sec:ecology}).}
\subsection{Normative assumptions for diversity}
\label{sec:philo}
\citet{sarkar_diversity_2010} examines diversity across ecological, biological, cultural,
and linguistic contexts from a philosophical perspective. He observes that
diversity emerges as a major value
of our times
and %
what all studied fields %
have in common is \textit{richness} (the number of
categories
in a system), a
``rather trivial concept of diversity'', according to
\citet[p.~136]{sarkar_diversity_2010}. What differs fundamentally across fields are the
justifications for endorsing diversity as a goal. They often rest on implicit normative assumptions, which have to be excavated.

For biodiversity, diversity is a normative goal by design, as the discipline was
created
to protect it. Two broad normative justifications are offered: (i)
\emph{intrinsic} value independent of all human considerations, and (ii)
\emph{anthropocentric} value, i.e.\ it serves human interests (e.g.\ food,
medicines). Since not all biological entities can be preserved, choices must be
made about which to protect. These choices rely on these normative assumptions.

In ecology, a major motivation for
endorsing diversity
is the hypothesis 
that
more diverse ecological communities are expected to be more stable than less
diverse ones (although counterexamples do exist).
The normative assumption is thus that
\emph{species extinction}
is not desirable while \emph{stability} is. 

The normative justification for protecting cultural and linguistic diversity
often mirrors biodiversity (e.g.\ preserving a ``pool of knowledge'' to respond
to environmental challenges). However, the
ultimate, often implicit,
normative assumption is 
respect for \emph{human rights}, which requires
protecting
cultural (and
linguistic) diversity.

Crucially, \citeauthor{sarkar_diversity_2010} argues that these normative assumptions
dictate the selection and adequacy of diversity measures. Different goals
require different ways of quantifying diversity.
It is therefore essential for NLP to consider \emph{why} diversity is of interest in
the field %
and thus make educated choices of diversity measures.
Here, we take steps towards answering this aspect.

\subsection{Stirling's cross-domain framework}
\label{sec:stirling}
Moving from normative questions to structural ones,
\citet{stirling_general_2007}
observes that while diversity is studied across radically different contexts,
the underlying properties remain remarkably similar across fields.
He proposes a cross-disciplinary unified
framework for characterizing diversity. Its basic premise is that diversity
is a property of a system whose \emph{elements} (also called \emph{options} or
\emph{items}) can be apportioned into \emph{categories} (also called
\emph{types}). For example, in ecology, \textit{categories} are typically
species and \textit{elements} are specimens.

Diversity measures that rely on this element/category dichotomy can then be positioned along three dimensions: \textbf{variety}, which focuses on the number of categories; \textbf{balance},
which considers the evenness of the distribution of elements in categories; and
\textbf{disparity}, which targets the extent of the differences between
categories.\footnote{The same three dimensions were first introduced by
\citet[p.427--428, Table 27]{stirling_power_1994} for the sake of economy, drawing inspiration from ecology.}
``All else being equal, the greater the variety, the greater the diversity'' \citep[p.~709]{stirling_general_2007}.
The same holds for balance and disparity. For instance, a community with 20 species has higher variety, and therefore higher diversity, than one with 10 species. %
A community with 3 species of 100 specimens
each is more balanced, and therefore more diverse, than one hosting 3 species
with 150, 100, and 50 specimens, respectively. 
A community with 5 bird species and 5 insect species has higher disparity than one with 10 insect species. 
\final{Similar considerations hold for other scientific domains. For instance, in economy, an electricity generating portfolio comprised of deep-mined hard coal, nuclear and wind power technologies is more disparate, and therefore more diverse, than one consisting of the same proportions of deep-mined hard coal, open-cast hard coal and brown coal plant \cite{Stirling-economics_98}.
Measures may tackle one or more dimensions at once.}\footnote{In the same paper, Stirling also proposes a concrete unified formula to encompass the three dimensions and cover diversity measures from different domains. This formula was later shown to have flaws
\citep{%
    rousseau_repeat_2018,%
    leydesdorff_interdisciplinarity_2019%
}, and we do not consider it here.}

\subsection{Diversity measurement in ecology}
\label{sec:ecology}

Ecology provides a particularly instructive case of how a field can move from the confusion of proliferating measures to greater clarity
by
convergence.

\paragraph{Proliferation of measures} Dozens of diversity measures were defined and applied to studies of various species and their habitats, including both nonparametric \citep{smith_consumers_1996}
and parametric \cite{MagurranAnneE2004Mbd} ones.\footnote{Parametric approaches, contrary to nonparametric ones, assume a particular species abundance distribution and use
its parameters
to estimate diversity \cite{chao_unifying_2014}.} In developing these measures, ecology borrowed extensively from information theory, through its use of parameterized entropies \citep{patil_diversity_1982} and
related transformations \citep{hill_diversity_1973}. Distance measures
were used to capture functional differences (e.g.\ body
features, behavior) and positions in the phylogenetic tree
\cite{mouchet_functional_2010}. 

\paragraph{The ``nonconcept'' problem} The proliferation of diversity measures soon became overwhelming. According to \citet[p.~577]{hurlbert_nonconcept_1971}, ``\emph{species diversity} has been defined in such various and disparate ways that it now conveys no information other than \emph{something to do with community structure}; species diversity has become a nonconcept''. Decades later, \citet{Ricotta2005ThroughTJ} still spoke of a ``jungle of biological diversity''.

\paragraph{Debates and convergence} The recognition of this problem led to productive debates. Some authors argued that diversity itself remains meaningful but %
measures should be carefully reviewed. 
Debates took place about desirable properties of diversity measures
\citep{smith_consumers_1996,jost_entropy_2006,hoffmann_is_2008,jost_mismeasuring_2009}\footnote{These
properties describe e.g.\ how diversity scores should behave when: (i) a
population is subdivided into two smaller ones, (ii) two smaller dissimilar
communities are merged into one, (iii) the abundance of a more frequent species
is increased at the expense of a less frequent one, etc.}  and a growing
consensus emerged around the use of \emph{\citeauthor{hill_diversity_1973} numbers} (\citeyear{hill_diversity_1973}), a family of diversity
measures that enable interpretability of diversity scores.%
\footnote{An example for the interpretability of diversity relies on the axiom that in a
perfectly balanced community,
diversity is equal to its number of species
\cite{jost_entropy_2006}. 
Refinements of this axiom assume that in such a maximally diverse
community, all species should also be totally dissimilar
\cite{leinster_measuring_2012}. For a
diversity measure $D$, and a
community $C$, to have interpretability, find a perfectly balanced community $C'$
such that $D(C')=D(C)$. The number of species in $C'$ is a pivot
to interpret various diversity measures.}
It can be shown
that many commonly used diversity measures can be seen as
special cases of a unified formula
\citep{renyi_measures_1961,patil_diversity_1982,stirling_general_2007}.
It was
also suggested that, rather than by unique scores, diversity should be
characterized by \emph{profiles} -- plotting the value of the unified formula
against its parameter, %
which gives more or less importance to different
dimensions of diversity \cite{leinster_measuring_2012,chao_unifying_2014}.

While debates continue, the field has achieved significant progress in
conceptualizing and measuring diversity. As a result,
diversity in ecology has become a rather mature and formalized topic with
substantial literature which we can draw from.

\section{Data and methodology}
\label{sec:methodology} 

Section~\ref{sec:other-fields} shows that diversity is valued across multiple fields, yet the normative assumptions and measurement frameworks vary considerably. In NLP research, while diversity is frequently invoked, there has been no systematic attempt to understand why it matters, how to define it, and how to measure it. 

To address this gap, 
we analyzed papers from the ACL
Anthology\footnote{\href{https://aclanthology.org/}{https://aclanthology.org/}} that include
``diverse'' or ``diversity'' in their title. As of July 26th, 2024, the ACL
Anthology hosted 97,466 papers; we focused on those from the last 5 years
(January 1st, 2019 to July 26th, 2024) with those terms in their title.\footnote{Sec.~\ref{section:discussion} discusses the consequences of this choice.}

This selection resulted in 308 papers,
which were manually analyzed. We filtered out 39 irrelevant
papers.\footnote{Papers judged irrelevant: (i) not written in English, (ii)
about non-linguistic diversity (e.g.\ biodiversity, neurodiversity), (iii)
``diversity'' in their title only because they cite the name of a
workshop/shared task or are volumes, (iv) slides, (v) ``diversity'' only in the
title and never in the body of the paper, (vi) describe projects, shared tasks
or event/thesis proposals.}
The
remaining 269 papers were annotated for
key aspects of how diversity is addressed within NLP research. Seven researchers participated in the annotation. To ensure annotation quality, adjudication was conducted for challenging cases. 

\section{Initial observations} 
\label{sec:initial-observations} 

Our preliminary review of the 269 papers
revealed several recurring issues in how diversity is conceptualized and measured.

First, diversity is generally assumed to be important in many tasks. High diversity 
is desirable,
while low diversity 
(e.g.\ due to training
models on synthetic data) %
is a concern. However, the motivations for valuing diversity are rarely explicit.  

Second, alongside the technical meaning of ``diversity'' as a shortcut for a particular quantifiable property, we witness frequent use of the term ``diverse'' in a common everyday sense, simply meaning \textit{different from one another}. This is the case in 82 papers.
While such uses are of course legitimate,\footnote{As
a reviewer
indicated,
a similar shift between common and technical meaning
concerns other terms like \emph{significant}.} they make the design of standard
vocabulary more challenging.  

Third, we observe cross-paper inconsistencies. Diversity measures vary
considerably across papers, and many are ad hoc or unclear, even after
conducting in-depth analysis or consulting the papers where they were first
defined. Many different aspects of NLP are subject to diversity quantification,
yet there is no uniform terminology and methodology, to the point of calling the
same measure different names or using the same name for different measures. 

Fourth, the choice of particular diversity measures is rarely justified and their properties are hardly addressed. 

These observations motivated our more systematic investigation of the 188 papers in which diversity is quantified (henceforth our corpus). 
\final{To this aim, we took inspiration from \citet{sarkar_diversity_2010} and \citet{stirling_general_2007} (Sec.~\ref{sec:other-fields}). Two main reasons underlie this choice.}

\final{First, these authors notably draw upon ecology, where diversity not only is a mature topic (Sec.~\ref{sec:ecology}) but where commonalities between animal species and world languages have been observed \cite{sole_diversity_2010}.}

\final{Second, \citeauthor{sarkar_diversity_2010}'s and \citeauthor{stirling_general_2007}'s frameworks are sufficiently high-level to embrace different scientific domains
and are therefore likely to apply to NLP as well.
Moreover, similarly to the fields studied by \citeauthor{sarkar_diversity_2010}, the motivations behind diversity endorsement in NLP are often implicit. NLP diversity scenarios frequently involve countable elements apportioned into categories (words into types, documents into languages, etc.), whose probability distributions are examined and whose distances are subject to aggregation. These are precisely the structural setups presupposed in \citeauthor{stirling_general_2007}'s framework.}

\final{Thus,} 
inspired by %
\citet{sarkar_diversity_2010} and \citet{stirling_general_2007}, we developed an analytical framework organized around four perspectives:
\emph{why} diversity is important
(Sec.~\ref{section:why}), \emph{what}
diversity is measured on
(Sec.~\ref{section:what-diversity-is-measured-on}), \emph{where} in the
research pipeline and NLP area diversity is quantified
(Sec.~\ref{section:where-diversity-is-measured}), and \emph{how} this
quantification is performed (Sec.~\ref{section:how-diversity-is-measured}).  
\final{This framework, and its enrooting in Sarkar's and Stirling's premises, indeed allowed us to better understand the diversity measurement landscape in NLP (Sec. \ref{section:discussion}) and to make recommendations for future developments in the field (Sec.~\ref{sec:recommandations}).  }

\section{\emph{Why} diversity is important in NLP}
\label{section:why}

A large majority of papers in our corpus endorse diversity but do not justify this stand. 
Following \citet{sarkar_diversity_2010}, we uncover the normative assumptions behind
the quest for diversity and group them into two families: ethical and practical.

\subsection{Ethical normative assumptions}
\label{sec:why-ethical}

Ethical normative assumptions are partly reminiscent of the respect for human rights (Sec.~\ref{sec:philo}). 

\paragraph{Equality and inclusiveness}
NLP should promote digital inclusiveness \cite{joshi-etal-2020-state}.
Datasets, systems, and benchmarks are expected to equally serve all users
\cite{khanuja-etal-2023-evaluating,liu-etal-2024-multilingual}, by representing
different languages, language families and scripts
\cite{kodner-etal-2022-sigmorphon,goldman-etal-2023-sigmorphon} and mitigating
the supremacy of English and English-centric bias
\cite{pouran-ben-veyseh-etal-2022-minion,asai-etal-2022-mia}. They should also
fairly account for diverse cultures
\cite{yin-etal-2021-broaden,mohamed-etal-2022-artelingo,keleg-magdy-2023-dlama,bhatia-shwartz-2023-gd,liu-etal-2024-multilingual},
human perspectives \cite{parrish-etal-2024-picture} and opinions
\cite{zhang-etal-2024-fair}. 

\paragraph{Protection of users}
Protection of users against misuse of technology should be ensured. For instance, pre-trained language models are fine-tuned with diverse prompts and responses, so as to limit the toxic and offensive content in LLM answers \cite{song-etal-2024-scaling}. More diverse attention vectors in a transformer make it less sensitive to adversarial attacks, which try to fool it, e.g.\ to misclassify sentiment or toxicity of a text \cite{yang-etal-2024-pad}.

\paragraph{Educational quality} When used in education, NLP should promote its high quality by covering a large variety of topics \cite{hadifar-etal-2023-diverse}. 

\paragraph{Methodological rigor}
NLP methods should align with NLP deontology.
Thus, more diverse benchmarks are created to make evaluation more reliable \cite{chen-etal-2023-unisumm}, notably because they show the out-of-domain performance 
\cite{pradhan-etal-2022-propbank}. As a result, the remaining challenges in NLP are better highlighted \cite{kim-etal-2023-aint}. Moreover, a dataset's diversity proves more critical in evaluation than its size \cite{miao-etal-2020-diverse}. 

\subsection{Practical normative assumptions}
\label{sec:why-practical}
Practical normative assumptions focus
on improving the functionality and performance of systems.

\paragraph{Meeting user expectations} 
One user expectation is informativeness and it was argued that more diverse generated text is less generic and more informative for users \cite{park-etal-2023-dive}. 
Since diversity is seen as an inherent property of human language, it has become a user expectation also towards machine-generated language. Thus, generation models should now operate in one-to-many scenarios, i.e.\ produce a diverse spectrum of outputs
\cite{li-etal-2016-diversity, kumar-etal-2019-submodular,liu-etal-2020-diverse,han-etal-2021-generating,shao-etal-2022-one,puranik-etal-2023-protege,e-etal-2023-divhsk,hwang-etal-2023-knowledge}
rather than a single most optimal output. Particularly high diversity expectations are encountered in dialog
\cite{lee-etal-2022-randomized}, where more diverse system reactions increase user engagement \cite{akasaki-kaji-2019-conversation,kim-etal-2023-persona}. Machine-generated text is also expected to be natural and the diversity of human language is often seen as an upper bound in this respect
\cite{schuz-etal-2021-diversity,cegin-etal-2023-chatgpt,liu-etal-2024-prefix}.

\paragraph{Improving performance} %
Expectedly,
increasing performance
of NLP tools is considered desirable. Therefore, authors increase the diversity of training data to
improve
performance in parsing %
\cite{liu-zeldes-2023-cant}, question answering
\cite{yadav-etal-2024-explicit}, semantic role labeling \cite{tripodi-etal-2021-united-srl}, solving math problems \cite{shen-etal-2022-seeking}, natural language generation 
\cite{zhang-etal-2021-trading,thompson-post-2020-paraphrase,
palumbo-etal-2020-semantic,li-etal-2021-textbox}.
Ensemble models are built of diverse submodels,
to achieve better performances than a single model \cite{song-etal-2021-deepblueai-semeval,kobayashi-etal-2022-diverse}. 
In classification, class names are extended with diverse semantically close keywords,
to increase accuracy \cite{yano-etal-2024-relevance}. This positive impact of diversity on performance stems from a better coverage %
of domains, tasks, genres, hallucinations types etc. \cite{xia-etal-2024-hallucination}.
\subsection{Normative tensions and value conflicts}
\label{sec:why-other}

While an implicit endorsement of diversity exists across the field, some papers
in our corpus do not share this view, but take more nuanced positions,
recognizing that diversity is not an absolute good but often exists in tension
with other salient objectives. These conflicts arise primarily when the pursuit
of diversity
begins to compromise the functional or ethical integrity of a system.

It is shown that simultaneously maintaining generative quality and 
diversity
is hard \cite{ma-etal-2024-volta}. This tension, often called the
\emph{quality/diversity trade-off}
\cite{ippolito-etal-2019-comparison,zhang-etal-2021-trading,shao-etal-2022-one},
or \emph{faithfulness/diversity trade-off} \cite{chen-etal-2023-fidelity}, shows
that diversity comes at a ``price''.  Consequently, some authors propose measures which assess both dimensions simultaneously \cite{narayan-etal-2022-well,alihosseini-etal-2019-jointly}.

Some works posit that diversity should be ad\-jus\-ted to the task
\cite{liu-etal-2024-self-regulated}. For instance, answers to factual questions
call for precision and determinism (low diversity), while storytelling requires
creativity and surprise (high diversity). This reflects
\quotecite{sarkar_diversity_2010}
warning that ``blanket
endorsement'' of diversity %
is unjustified.

Moreover, the closeness (in diversity) of generated text to human production might not be an objective per se. 
For instance, the difference in diversity patterns between human-generated and AI-generated text may help distinguish these two categories. 
This serves %
bot detection on social media, avoidance of fake news, and protection of democracy \cite{kosmajac-keselj-2019-twitter}.

\subsection{Understanding diversity}

One final reason for the interest in diversity 
is to theorize it, and systematize its measurement, so as to understand better its nature and implications \cite{ploeger-etal-2024-typological}, make educated choices of diversity
measures \cite{lai-etal-2020-diversity,tevet-berant-2021-evaluating,lion-bouton-etal-2022-evaluating}, and offer a reliable comparative framework for NLP \cite{poelman-etal-2024-call}. Our work falls precisely within this motivation.

\section{\emph{What} objects are measured for diversity}
\label{section:what-diversity-is-measured-on}

The papers in our corpus address a wide range of ``diversities'' across NLP areas. We take inspiration from \citet[p.~709]{stirling_general_2007}, who noted that it is when the elements and categories are identified that similarities and particularities of various understandings of diversity become clear. Broadly speaking, the categories that emerge from our analysis of the corpus fall into the three following classes.

\subsection{In-text categories}\label{sec:in-text}
In-text categories are 
associated with linguistic properties that are implicitly
assumed to be inherent to a text, whether it is written, spoken, or signed. They are addressed and quantified in 119 papers (out of 188) in our dataset. For
example, \citet{guo-etal-2024-curious} find that
through iterations of training a model with its predecessor's synthetic output,
texts exhibit a consistent decrease in their lexical, semantic and
syntactic diversity. Lexical diversity is operationalized in their system with
(i)
words as elements and unique words as
categories and (ii) \emph{n}-grams as elements and unique
\emph{n}-grams as categories. Semantic and syntactic diversity in their
approach can be understood as \emph{disparity}, 
with sentences being both
categories and elements. Similarly, sets of LM-generated responses
\citep{gao-etal-2019-jointly,
tevet-berant-2021-evaluating,han-etal-2022-measuring}, captions
\citep{schuz-etal-2021-diversity}, translations
\citep{burchell-birch-and-kenneth-heafield-2022-exploring, shao-etal-2022-one,
wu-etal-2020-generating}, paraphrases \citep{kumar-etal-2019-submodular,
bawden-etal-2020-study, cao-wan-2020-divgan} are evaluated
for
the
diversity of the words or word sequences that they contain, as well as the
diversity of their semantic contents and syntactic structures.
More rare but interesting instances of in-text diversity are
when
categories are not defined beforehand but emerge automatically e.g.\ by clustering \cite{han-etal-2022-measuring}.

\subsection{Meta-linguistic categories}\label{sec:meta-ling}
Meta-linguistic
categories relate not to texts themselves, but to the classifications of
texts within datasets.
We found them
in 51 papers.
Multilingual datasets are often evaluated
based on
the languages they
contain. In the simplest case, languages are categories. 
Diversity is characterized by \emph{variety}, the number of
languages, and \emph{disparity}, the distinctive typological or phylogenetic
properties of each language \citep{pouran-ben-veyseh-etal-2022-minion,
sarti-etal-2022-divemt}. \emph{Balance} becomes relevant when the elements within each category are texts or tokens.

In a more nuanced characterization of typological
diversity, languages are classified according to language families and,
sometimes, branches within them \citep{longpre-etal-2021-mkqa,
kodner-etal-2022-sigmorphon, kumar-etal-2022-indicnlg}. In our terminology,
languages in that case are the elements and the language branches or
families, the categories. However, although the term `typological
diversity' is often evoked, there is a real need for adopting a systematic
approach for measuring and comparing it
\citep{ponti-etal-2019-linguistic-typology, poelman-etal-2024-call}. 

Languages are categories for \citet{dunn-etal-2020-measuring}, who draw on the
meta-linguistic diversity of Twitter data as a means for understanding the
linguistic landscape of countries. With tweets as elements categorized
into the languages in which they were produced, the authors measure for each
country its linguistic diversity over time, and in particular the effect of
travel restrictions caused by COVID-19. 

Other text classifications evaluated
with regard to diversity are genres \citep{liu-zeldes-2023-cant} and domains or
time periods to which the texts belong
\citep{bhattacharyya-etal-2023-vacaspati}. In such cases, the elements
are linguistic items, mostly tokens, 
and the categories are the
meta-linguistic categories they belong to. There are also papers that target the
diversity of text producers, e.g.\ the racial identity of language signers
\citep{gueuwou-etal-2023-jwsign} or the political opinions of authors
\citep{zhang-etal-2024-fair}.

\subsection{Processing categories}\label{sec:processing}
A third type of categories, addressed only in
24 papers, are associated with
the processes applied to the data. Thus, processing categories are external to the text itself. This is a wide
class; we find diversity ascribed to annotations
\citep{weerasooriya-etal-2023-disagreement},
annotators
\citep{parrish-etal-2024-picture,creanga-dinu-2024-designing}, to the (ensemble
of) models
trained
\citep{greco-etal-2022-small,
kobayashi-etal-2022-diverse}, to the types of tasks performed
\citep{qiu-etal-2021-different, zhang-etal-2024-fine}, and to the attention
vectors in the
design of a model \citep{huang-etal-2019-multi,
yang-etal-2024-pad}. In most of these cases there are no higher-order
classifications, i.e., \emph{categories} are \emph{elements}. One exception is
\citet{yang-wan-2022-investigating}, who investigate metric diversity for long
document summarization: the \emph{elements} here are evaluation metrics (e.g.\
BLEU,
SPICE) which are classified into \emph{categories} (e.g.\
translation, summarization, semantics).

\section{\emph{Where} diversity is measured}
\label{section:where-diversity-is-measured}
We analyzed where diversity is quantified in two ways. For a broad overview
of whether some NLP subfields are more concerned with diversity than others, we
examined the NLP area of each paper. We also determined at which stage of a
standard machine learning pipeline quantification happens.

\subsection{NLP areas}
We annotated each paper in our
corpus according to the main NLP areas defined in
ARR\footnote{\href{https://aclrollingreview.org/cfp}{https://aclrollingreview.org/cfp}.
The ARR area the paper was submitted to was unavailable,
but we used the one in the title if
present and our best guess if not.} that the study targets, as shown in
Table~\ref{table:datasets_sorted_desc}.
Given the recent uptake in NLP for generating data, it is not surprising that one of the most frequent areas is
\textit{Generation} (27 papers, thus 14\% of our corpus). All papers in that area aim
at generating outputs that are ``diverse'' (mostly in terms of in-text
diversity, Sec.~\ref{sec:in-text}). This also holds for the areas of
\textit{Dialogue}, \textit{Machine Translation}, \textit{NLP Applications},
\textit{Question Answering} and \textit{Summarization}, in which text is often
generated. \textit{Resources and Evaluation} is also an area in which diversity, mostly
meta-linguistic diversity (Sec.~\ref{sec:meta-ling}), is prominently used (27
papers, 14\% of the total number of papers). Concerns about diversity are
also apparent in \textit{Machine Learning for NLP} (31 papers, 16\% of the total number of papers) which encompasses classification tasks and algorithms to obtain diverse generated data. %

\begin{table}%
\small
\centering
\begin{tabular}{p{0.84\linewidth}r}
\toprule
\textbf{ARR area}                     & \textbf{\#} \\ \midrule

Dialogue \& Interactive Systems & 19 \\
Discourse \& Pragmatics & 1 \\
Generation & 27 \\
Information Extraction & 1 \\
Language Modeling & 8\\
Machine Learning for NLP & 31 \\
Machine Translation & 11 \\
Multilingualism and Cross-lingual NLP & 11\\
Multimodality, Grounding to Vision & 6 \\ %
NLP Applications & 16\\
Phonology, Morphology, Segmentation &  4\\ %
Question Answering & 6 \\
Resources and Evaluation & 27 \\
Semantics: Lexical and Sentence-level & 8 \\
Summarization & 8 \\
Speech Recognition, Text-to-speech & 2 \\ %
Syntax & 2\\  %
\midrule
Computational Social Sciences;
Efficient/Low-Resource Methods;
Ethics, Bias \& Fairness;
Human-centered NLP;
Information Retrieval \& Text Mining;
Interpretability \& Analysis of Models;
Linguistic theories \& Cognitive Modeling;
Sentiment \& Stylistic Analysis & 0 \\

\bottomrule
\end{tabular}
\caption{Number of papers per ARR area in our corpus.\label{table:datasets_sorted_desc}} \end{table}

\subsection{Pipeline stages} 
\label{sec:pipeline}

Figure~\ref{fig:pipeline-stages} shows the
different stages of a standard machine learning pipeline, detailed here. %

\begin{figure}
\begin{center}
\begin{tikzpicture}
    \node[draw, inner sep=-0.0cm, fill=blue!10] (datacollection) at (0.0, 0.0) {\begin{tabular}{c} Data \\ collection \\ \textbf{(37, *1)} \end{tabular}};
    \node[draw, inner sep=-0.0cm, fill=blue!10] (annotation) at (0.0, 1.5) {\begin{tabular}{c} Annotation  \\ \textbf{(2)} \end{tabular}};
    \node[draw, inner sep=-0.1cm, fill=green!10, ellipse] (input) at (2.5, 1.5) {\begin{tabular}{c} Input \\ \textbf{(25, *5)} \end{tabular}};
    \node[draw, inner sep=-0.0cm, fill=blue!10] (systemconstruction) at (2.5, 0.0) {\begin{tabular}{c} System \\ construction \\ \textbf{(21, *4)} \end{tabular}};
    \node[draw, inner sep=-0.1cm, fill=green!10, ellipse] (output) at (5.0, 1.5) {\begin{tabular}{c} Output data  \\ \textbf{(86, *8)} \end{tabular}};
    \node[draw, inner sep=-0.0cm, fill=blue!10] (evaluation) at (5.0, 0.0) {\begin{tabular}{c} Evaluation  \\ \textbf{(4)} \end{tabular}};
    
    \draw[->] (datacollection) -- (input);
    \draw[->] (input) -- (systemconstruction);
    \draw[->] (systemconstruction) -- (output);
    \draw[->] (output) -- (evaluation);
    \draw[dashed] (annotation) -- (input);
\end{tikzpicture}
\end{center}
\caption{High-level stages in standard machine learning pipelines (rectangles for
processes, ellipses for states), with numbers
of papers quantifying diversity. Papers with two-stage quantification indicated with *.\label{fig:pipeline-stages}}
\end{figure}

\paragraph{Data collection} Different methods or criteria are used to select data with diversity
in mind. Elements are thus methods/criteria and categories are groupings of these. \citet{kim-etal-2023-aint} focus on math word problems and
create a more diverse dataset than previously existing ones, in terms of problem
types (arithmetic, correspondence, comparison, geometry, possibility), lexical
usage (as measured by a lexical diversity measure), languages (they include both
English and Korean), and intermediate solution forms (different equation
templates).

\paragraph{Annotation process} The annotation is done with diverse
pools of annotators (human or machine). Elements are the annotators, categories are some annotator properties (e.g.\ perspective,
sociodemographics). In NLI for example, \citet{creanga-dinu-2024-designing}
consider the diversity of human annotators, with annotators as \textit{elements}
and perspectives as \textit{categories}, where diversity pertains thus to
processing (Sec.~\ref{sec:processing}). The potential benefits of such an
approach are to better represent different worldviews, and consequently improve
system performance.
\paragraph{Input data} The data given to the system is
diverse. Elements are often
occurrences of linguistic entities and
categories are in-text or meta-linguistic groupings of these elements (e.g.\ token per lemma,
document per genre).
An example by \citet{liu-zeldes-2023-cant} shows
that the meta-linguistic diversity
(Sec.~\ref{sec:meta-ling}) of training sets positively impacts
downstream performance in discourse parsing.
Diversity is operationalized
based on
the number of genres
in the training sets
(e.g.\ academic,
bio, interview) and the number of discourse units
per
genre.
\paragraph{System construction} The process itself is diversity-driven. For
instance, in a standard machine learning process, the loss function can be a
diversity measure. \citet{yang-etal-2024-pad} assume that if a transformer-based
model has diverse patterns in its attention layers, it is less  sensitive to
attacks by adversarial examples. Thus, while training the model, the loss
function is defined as the volume of the geometry formed by attention vectors,
and maximization of this volume is sought. Here, attention vectors are elements and their own categories.

\paragraph{Output data}
System's output is measured for diversity. For instance,
\citet{han-etal-2022-measuring} propose a semantic diversity measure of
generated answers, which correlates better with human judgments on answer
diversity. Elements are response instances, and categories semantic clusters.

\paragraph{Evaluation} Different metrics/criteria are used to evaluate systems' outputs, for instance to reinforce their applicability. Elements are metrics/criteria and categories are
groupings of these. This falls under diversity of processing (Sec.~\ref{sec:processing}) or meta-linguistic diversity (Sec.~\ref{sec:meta-ling}). For instance, \citet{yang-wan-2022-investigating} use multiple metrics drawn from multiple NLP tasks to evaluate summaries, while \citet{bhatia-shwartz-2023-gd} employ human evaluators from 5 different cultures to judge efficiency of inference models. 

\paragraph{In which
stage(s) is diversity quantified?} For each of the 188 papers in our corpus, we identified the stage(s) in which the quantification of
diversity occurs. 183 papers fit the pipeline (some survey or opinion papers
do
not). 
Figure~\ref{fig:pipeline-stages} gives, for each stage, the number of papers in which diversity is quantified in that stage (starred * if 
quantification happens in two stages).  
``Output data'' is unsurprisingly the stage
in which diversity is most often quantified
(86 out of 188 papers):
21 of these are from \textit{Machine Learning}, 19 from \textit{Generation}, 13 from \textit{Dialogue}, 9 from
\textit{NLP Applications} and 5 from \textit{Machine Translation}. The ``Data
collection'' stage is also prominent, with 15 of the
27
papers belonging to the \emph{Resources and Evaluation} area. In 9 papers, diversity is
measured in two stages: for 4 papers, at Input data \& Output
data; for 3, at System construction \& Output data; for 1, at Data collection \& Output data; and in another 1, at Input data \& System construction.

\final{The distribution of pipeline stages discussed above is computed over our entire corpus. Broken down by year of publication (Figure~\ref{fig:diversity-evolution-stage}), the values remain largely stable, with notable increases in Input and Data collection in 2022, as well as in Data collection and Output data in 2023.}

\begin{figure}
    \centering
    \resizebox{\linewidth}{!}{
    \input{images/diversity-evolution-time-stage.pgf}
    }
    \caption{Evolution in time of pipeline stages in which diversity is quantified. Note that the data for 2024 ends on July 26th.
    \label{fig:diversity-evolution-stage}}
\end{figure}

\section{\emph{How} diversity is measured}
\label{section:how-diversity-is-measured}

Looking at all 188 papers
in our corpus, 
we found over
100 different names of diversity measures. Most are used in a handful of papers,
which makes straightforward comparisons hard. However, when applying Stirling's
cross-domain framework (Sec.~\ref{sec:stirling}), we can abstract away from
categories (Sec.~\ref{section:what-diversity-is-measured-on}), elements and
distance measures (if any) and see a high-level picture emerge.

The outcome of this effort is shown in Tables~\ref{tab:conforming-measures}-\ref{tab:non-conforming-measures}, with a split into measures which do and do not fit \citeauthor{stirling_general_2007}'s framework, respectively.
Each bullet point lists different names found for the same measure, with sample papers using them. Measures %
based on similar mechanisms are grouped into families, as summarized below. 

Henceforth, the set for which diversity is measured will be called the \emph{observed set}. Let $n$ and $m$ be the number of its categories and elements (called \emph{observed categories} and \emph{observed
elements}), respectively. Let $P = \left<p_1,...,p_n\right>$ denote the array
of category frequencies, and $D =
\left<\left<d_{1,1},...,d_{1,n}\right>,...,\left<d_{n,1},...,d_{n,n}\right>\right>$
the matrix of distances between the categories.

\subsection{Conforming measures}
\label{sec:conforming}

Over 40 diversity measures, bearing over 60 different names, and found in 167 papers from our corpus, could successfully be mapped on Stirling's framework, which made 7 families emerge.

\paragraph{dRichness}
This family contains two variants of \texttt{richness}, i.e. simply the number of categories,
which is the most elementary variety measure:
\begin{equation}
\text{\texttt{richness}}\left(n, m, P, D\right) = n
\end{equation}
Most papers found here do not explicitly mention any measure, but simply state
the number of
categories. They deal most often with meta-linguistic 
diversity (Sec.~\ref{sec:meta-ling}), e.g. the number of languages, but
sometimes also with in-text diversity, e.g. number of
distinct n-grams (\texttt{Dist-n}).

\paragraph{dBalance}

This family gathers pure balance measures. Only 2 examples were found but we
judge them important enough to be singled out. The \texttt{E$_{2,1}$ evenness}
draws upon Hill numbers highly valued in the ecological SOTA 
(Sec.~\ref{sec:ecology}). The \texttt{Zipfian coefficient} addresses the specificity of distributions in language data.

\paragraph{dEntropy}
The main measure here is Shannon-Wiener \citeyearpar{shannon_mathematical_1948,wiener_ergodic_1939} \texttt{entropy}, i.e. the weighted average of the surprise $-\ln\left(p_i\right)$ 
of observing a
category with index $i$:
\begin{equation}
\text{\texttt{entropy}}\left(n,m,P,D\right) = -\sum\limits_{i=1}^{n} p_i
\ln\left(p_i\right)
\end{equation}
It is monotonic to $n$, increases with evenness and its maximum $\ln\left(n\right)$ is reached for a uniform distribution. Therefore it is considered a hybrid between variety and balance. It applies to the distribution of n-grams, to categories constructed automatically by clustering (\texttt{sem-ent}) or to language features (\texttt{typological index}).

\paragraph{dVBOth}

Here, we group 3 variety-balance measures distinct from entropies. They are found in only 3 papers, but draw from other fields.

\paragraph{dAvePairwise}

These measures aggregate the distance matrix %
and can be cast as disparity.
There
are many
measure names due to different categories (sentences, words), their representations (strings, syntax trees, vectors, kernels) and distance/similarity measures used.
But they all
boil down
to the \texttt{pairwise} average distance between categories (higher is more diverse):
\begin{equation}\label{eq:pairwise}
\text{\texttt{pairwise}}\left(n, m, P, D\right) =
\frac{2*\sum\limits_{i=1}^{n}\sum\limits_{j=1}^{i-1}d_{i,j}}{n(n-1)}
\end{equation}
or its complement: pairwise average similarity (lower is more diverse).\footnote{Note the $O(n^2)$ complexity.} 
Many papers from this family use measures relying on BLEU \cite{papineni-etal-2002-bleu}, initially designed as a performance measure in machine translation, and
repurposed
here as the complement of diversity. Notably, \texttt{self-BLEU} \citep{DBLP:conf/sigir/ZhuLZGZWY18} is the average of BLEU between each pair of utterances. This  follows the same logic as \texttt{pairwise} in (\ref{eq:pairwise}), with BLEU as the similarity function (lower is better), $n^2$ instead of $n(n-1)/2$ as a denominator, \final{and $j$ going from $1$ to $n$, because BLEU is not symmetric}. Variants of \texttt{self-BLEU} stem from restricting the length of n-grams compared by BLEU (\texttt{self-BLEU-3/4}), complementing BLEU to transform it into a distance (\texttt{BLEU-based discrepancy}), etc. 
Other underlying distance measures include cosine similarity, BERTScore \cite{bert-score}, token overlap, entailment score predicted by an NLI model, etc.
\paragraph{dVolume}

This is another small but interesting family, with aggregation of the distances (disparity) operationalized by the volume (possibly squared) of the geometry formed by vectors representing the categories or its main parameter (radius).

\paragraph{dDistOth}

This family gathers other instances of aggregations (disparities) such as (i) the sum, rather than the average, of pairwise distances (\texttt{max dispersion}, \texttt{NLI diversity}), (ii) the entropy of distances between vectors (\texttt{graph entropy}), (iii) 
the most distant categories (\texttt{corpus lexicon diversity}), etc.

\subsection{Non-conforming measures}
\label{sec:non-conforming}
The 42 remaining measures, bearing 47 names and found in 88 papers, have no counterparts in \citeauthor{stirling_general_2007}'s framework. We grouped them into 6 families.

The first three tackle relative rather than absolute quantification, i.e. they consider \emph{two} sets: a reference set $R$ and the observed set $O$, whose
diversity is estimated by comparison to $R$. In one version of this scenario, $R$
is considered diverse, e.g.\ it is curated with diversity in mind, and $O$
should be as close as possible to $R$. In another version, $O$ is expected to differ from $R$, e.g.\ generated
utterances should be different from the training utterances. 

\paragraph{dRelOverlap}
These measures focus on how many categories are shared by $R$ and $O$. One example is the \texttt{Jaccard} measure:
\begin{equation}
\text{\texttt{Jaccard}}\left(R,O\right) = \frac{
\left\vert \overline{R} \cap \overline{O} \right\vert
}{
\left\vert \overline{R} \cup \overline{O} \right\vert
}
\end{equation}
where $\overline{R}$ and $\overline{O}$ are the sets of categories in $R$ and $O$. It tackles the ratio of shared categories %
and has the range $[0,1]$. 
Other measures in this family include the \texttt{percentage of  categories} in $R$ which do or do not occur in $O$, or the \texttt{F1} \texttt{score}.

\paragraph{dRelDistrib}
These measures compare the reference and observed distributions of
$R$'s and $O$'s elements in categories, respectively. One example is
\texttt{Jensen-Shannon Divergence}: 
\begin{equation}
    \text{{JSD}}(Q\parallel P)=\frac{1}{2}D(Q\parallel M)+\frac{1}{2}D(P\parallel M)
\end{equation}
where $Q$ is the reference distribution, $P$ is the observed distribution,
$M=\frac{1}{2}\left(Q+P\right)$ is a mixture distribution, and $D\left(Q \parallel M\right)$ is Kullback-Leibler divergence. Other measures in this family include cross-entropy, area under curve, etc.

\paragraph{dRelDist}
Here, distances, e.g. BLEU, between the categories from $O$ and from $R$ are aggregated, which yields a distance between $O$ and $R$.

\noindent\paragraph{}The 3 remaining non-conforming  families concern absolute quantification, like those from Tab.~\ref{tab:conforming-measures}.

\paragraph{dTTR}
This family consists of variants of
\begin{equation}
\text{\texttt{type-token-ratio}}\left(n, m, P, D\right) = \frac{n}{m}
\end{equation}
which is the number of categories normalized by dataset size. A frequent variant is \texttt{distinct-n}, also called \texttt{dist-n} or \texttt{diverse-n}, which is the ratio of distinct n-grams to the total number of tokens, rather than to the number of n-grams.
\texttt{Distinct-n} is widely used in NLP, but
it has issues \cite{BestgenYves2024MLDi}. Importantly for us, it is not monotonic to the number of categories, which questions its place in the variety dimension. %
Other variants,
such as \texttt{MTLD} and \texttt{MATTR}, are TTR averages over sliding windows of variable or fixed sizes,
making
them insensitive to text length.

\paragraph{dHuman}

This family builds upon the assumption that humans are capable of judging diversity in text by introspection. 
Human annotators are presented with text samples that they have to \texttt{rank} for diversity or score along a diversity \texttt{scale}. We cannot know a priori whether humans rely on categories and elements for their judgment; therefore, this family cannot be cast in \citeauthor{stirling_general_2007}'s framework.

\paragraph{dOther}

This family gathers
measures that seem to fit neither
\citeauthor{stirling_general_2007}'s
framework,
nor
any of the previously defined families. They are mostly quite
specific to particular tasks (e.g. ensemble classification, sentiment or coreference resolution) and and do not seem to fit \citeauthor{stirling_general_2007}'s dimensions \final{or call for further investigation}. %

\paragraph{} Several global observations can be made.  
The most frequently used measures are those from dRichness (78 papers), dAvePairwise (52) and dTTR (43), with the first two being pure richness and pure disparity, respectively. There are only 3 cases of pure balance. Variety-balance hybrids are relatively popular, mainly due to entropy, but we see no hybrids with disparity.\footnote{Hybrids with disparity do exist in ecology
\cite{ricotta_towards_2006,leinster_measuring_2012,scheiner_metric_2012,chao_unifying_2014}.} There are many cases where the same measure bears different names (listed in the same bullet point) and it happens that the same name is used for different measures (\texttt{dist-n}, \texttt{self-BLEU}). Finally, in over 20 cases relative quantification is used.

\begin{table}[ht!]

  \begin{scriptsize}
  \centering
  
  \setlength{\tabcolsep}{0.8mm} %
  \renewcommand{\arraystretch}{0.4} %
  \setlength{\extrarowheight}{0pt}
  \setlength{\parskip}{0pt}

  \begin{tabular}{|p{0.025\linewidth}|p{0.925\linewidth}|} \hline
  $D$ & NLP diversity measures, conforming to \citet{stirling_general_2007}
  \\ \hline \hline
  \rotatebox[origin=r]{90}{\textbf{Variety}}
  &
  \textbf{dRichness} (78) \par
    \textbullet~number of categories \cite{vylomova-etal-2020-sigmorphon},
    Dist-n and Dist-S \cite{park-etal-2023-dive}, vocabulary size \cite{cegin-etal-2023-chatgpt}, number of unique words \cite{panagiaris-etal-2020-improving}, richness \cite{lion-bouton-etal-2022-evaluating} \par
    \textbullet~ave. \# categories per text \cite{ye-etal-2021-one2set,byrne-etal-2019-taskmaster} %
  \\ \hline
  \rotatebox[origin=r]{90}{\textbf{Balance}} 
  &
  \textbf{dBalance} (3) \par
  \textbullet~Hill-based E$_{2,1}$ evenness \cite{lion-bouton-etal-2022-evaluating} \par
  \textbullet~Zipfian coefficient \cite{hong-etal-2020-diverse,zhang-etal-2023-lingxi}
  \\ \hline
  \multirow{5}{*}{\rotatebox[origin=r]{90}{\textbf{Variety-balance}}}
  &
  \textbf{dEntropy} (21) \par
  \textbullet~entropy \cite{joshi-etal-2020-state,keleg-magdy-2023-dlama} \par
  \textbullet~sem-ent \cite{han-etal-2022-measuring} \par %
  \textbullet~syntax/morphology typological index \cite{samardzic-etal-2024-measure} %
  \\\cline{2-2}

  &
  \textbf{dVBOth} (3) \par
  \textbullet~Herfindahl-Herschman index \cite{dunn-etal-2020-measuring} \par
  \textbullet~Gini-index \cite{raza-etal-2022-accuracy} \par
  \textbullet~Simpson’s D index \cite{kosmajac-keselj-2019-twitter} 
  \\ \hline
  \multirow{10}{*}{\rotatebox[origin=r]{91}{\textbf{Disparity}}}
  &
  \textbf{dAvePairwise} (52) \par %
    \textbullet~Self-BLEU \cite{han-etal-2021-generating}, 
    pairwise BLEU \cite{e-etal-2023-divhsk}, 
    average pairwise BLEU \cite{shao-etal-2022-one}, 100-BLEU \cite{awasthi-etal-2022-diverse}, BLEU \cite{aji-etal-2021-paracotta} \par %
    \textbullet~BLEU-based discrepancy \cite{shu-etal-2019-generating}, 1-BLEU \cite{hu-etal-2019-large} \par
    \textbullet~average BLEU-based discrepancy \cite{shu-etal-2019-generating} \par%
    \textbullet~i-BLEU \cite{burchell-birch-and-kenneth-heafield-2022-exploring} \par
    \textbullet~Self-BLEU-n \cite{hwang-etal-2023-knowledge}, BLEU-n \cite{weir-etal-2020-cod3s} \par 
    \textbullet~ave. pairwise cosine \cite{kim-etal-2024-improving}, ADC \cite{hadifar-etal-2023-diverse}, average USE similarity \cite{dopierre-etal-2021-protaugment}, or dispersion of sentence embeddings \cite{guo-etal-2024-curious} \par %
    \textbullet~average syntactic lang2vec \cite{poelman-etal-2024-call} \par
    \textbullet~self-entailment \cite{narayan-etal-2022-well} \par %
    \textbullet~average tree edit distance \cite{hu-etal-2019-large,cegin-etal-2023-chatgpt}, parse tree edit distance \cite{jegadeesan-etal-2021-improving} \par
    \textbullet~mean tree kernel diff. \cite{burchell-birch-and-kenneth-heafield-2022-exploring}, ave. graph kernel dist. \cite{guo-etal-2024-curious}, DStree \cite{bawden-etal-2020-study} \par %
    \textbullet~i-chrF (character n-gram F-score) \cite{burchell-birch-and-kenneth-heafield-2022-exploring} \par
    \textbullet~pairwise average SBERT dissimilarity  \cite{weir-etal-2020-cod3s} \par
    \textbullet~Self-BERTScore \cite{narayan-etal-2022-well} \par %
    \textbullet~Self-BLEURT \cite{kwon-etal-2021-toward} \par %
    \textbullet~average of unique words in common \cite{bawden-etal-2020-study} \par
    \textbullet~average word-level Jaccard \cite{aji-etal-2021-paracotta}, mean Jaccard n-gram similarity \cite{li-etal-2022-evade}, or average intersection over union \cite{jegadeesan-etal-2021-improving} \par
    \textbullet~average pairwise lexical token overlap \cite{yadav-etal-2024-explicit} \par
    \textbullet~average reverse Jaccard of n-grams \cite{larson-etal-2019-outlier} \par
    \textbullet~average word-error rate \cite{jegadeesan-etal-2021-improving} \par
    \textbullet~inverse rank-biased overlap \cite{li-etal-2023-diversity} \par
    \textbullet~gate diversity \cite{lai-etal-2020-event}
  \\ \cline{2-2}
  &
  \textbf{dVolume} (3) \par %
    \textbullet~mean of ellipsoid radii across all axes \cite{lai-etal-2020-diversity} \par %
    \textbullet~squared volume of vectors' geometry  \cite{yang-etal-2024-pad} \par
    \textbullet~convex hull volume \cite{yu-etal-2022-data} 
  \\ \cline{2-2}
  &
  \textbf{dDistOth} (7) \par %
  \textbullet~graph entropy \cite{yu-etal-2022-data} \par
  \textbullet~max dispersion \cite{yu-etal-2022-data} \par %
  \textbullet~corpus lexicon diversity \cite{miao-etal-2020-diverse} \par
  \textbullet~intra-batch average/minimal distance  \cite{shi-etal-2021-diversity}, 
  average maximum (Jaccard) similarity \cite{hu-etal-2022-recurrence} \par
  \textbullet~average outlier \cite{stasaski-etal-2020-diverse} \par
  \textbullet~NLI diversity \cite{stasaski-hearst-2022-semantic} \par %
  \textbullet~triplet correlation loss function \cite{huang-etal-2019-multi}
  \\ \hline

  \end{tabular}
  \caption{\citeauthor{stirling_general_2007}'s dimensions ($D$) and NLP families (in bold) of diversity measures and sample articles using them.
  Number of papers per family in parentheses.\label{tab:conforming-measures}}
  \end{scriptsize}
\end{table}

\begin{table}[ht!]

\begin{scriptsize}
\centering

\setlength{\tabcolsep}{0.8mm} %
  \renewcommand{\arraystretch}{0.4} %
  \setlength{\extrarowheight}{0pt}
  \setlength{\parskip}{0pt}

\begin{tabular}{|p{0.95\linewidth}|} \hline
NLP diversity measures, non-conforming to \citet{stirling_general_2007}
\\ \hline \hline

\textbf{dRelOverlap} (9) \par %
   \textbullet~MS-Jaccard \cite{alihosseini-etal-2019-jointly} \par 
   \textbullet~Jaccard minmax \cite{samardzic-etal-2024-measure} \par
   \textbullet~\% novel \cite{schuz-etal-2021-diversity,park-etal-2023-dive} \par %
   \textbullet~vocabulary usage \cite{liu-etal-2024-prefix}, \% coverage \cite{schuz-etal-2021-diversity} \par %
   \textbullet~number of novel \cite{murahari-etal-2019-improving} \par
   \textbullet~overlap \cite{zhou-etal-2021-learning} \par %
   \textbullet~proportion of times a ground truth token was found in a model's nucleus \cite{sultan-etal-2020-importance} \par
   \textbullet~F1 between model-generated categories in the reference and in the predictions \cite{huang-etal-2023-examining}
\\ \hline
\textbf{dRelDistrib} (4) \par %
   \textbullet~Fréchet BERT Distance \cite{alihosseini-etal-2019-jointly} \par
   \textbullet~Bhattacharyya \cite{alihosseini-etal-2019-jointly} \par
   \textbullet~Jensen-Shannon Divergence \cite{xia-etal-2024-hallucination} \par
   \textbullet~data cross-entropy \cite{kumar-etal-2022-diversity} \par
   \textbullet~average frequency rank with respect to the training set  \cite{schuz-etal-2021-diversity}%
\\ \hline
\textbf{dRelDist} (8) \par %
   \textbullet~Self-BLEU \cite{cao-wan-2020-divgan} \par%
   \textbullet~recall \cite{gao-etal-2019-jointly} \par%
   \textbullet~Max-BLEU \cite{tu-etal-2019-generating}  \par
   \textbullet~Senti$_{MRR}$ \cite{wu-etal-2020-sentirec}  \par
   \textbullet~Senti@5 \cite{wu-etal-2020-sentirec}  \par
   \textbullet~Senti@10 \cite{wu-etal-2020-sentirec}
\\ \hline\hline

\textbf{dTTR} (43) \par %
\textbullet~Distinct-n \cite{ippolito-etal-2019-comparison},
Dist-n \cite{liu-etal-2024-prefix}, %
topic uniqueness \cite{li-etal-2023-diversity} \par
\textbullet~TTRn \cite{schuz-etal-2021-diversity}, proportion of distinct n-grams \cite{nishida-etal-2024-generating,song-etal-2024-scaling} \par
\textbullet~Measure of Textual Lexical Diversity \cite{fu-nederhof-2021-automatic,lee-etal-2022-randomized} \par
\textbullet~Moving-Average Type–Token Ratio \cite{lee-etal-2022-randomized,wu-etal-2024-lamini}, Mean Segmental Type-Token Ratio \cite{chang-etal-2022-programmable} \par
\textbullet~product of n-gram repetition proportions \cite{chen-etal-2023-fidelity} \par
\textbullet~proportion of duplicates \cite{bahuleyan-el-asri-2020-diverse}%
\\ \hline
\textbf{dHuman} (12) \par %
\textbullet~ranking \cite{eo-etal-2023-towards,park-etal-2023-dive,lahoti-etal-2023-improving,lee-etal-2023-partially} \par
\textbullet~scale \cite{diao-etal-2021-tilgan,narayan-etal-2022-well,kim-etal-2023-concept,yoon-bak-2023-diversity}
\\ \hline
\textbf{dOther} (12) \par %
\textbullet~Mean IDF \cite{stasaski-etal-2020-diverse} \par
\textbullet~max-gap \cite{lahoti-etal-2023-improving} \par %
\textbullet~average Q statistic \cite{kobayashi-etal-2022-diverse} \par
\textbullet~ratio errors \cite{kobayashi-etal-2022-diverse} \par
\textbullet~negative double fault \cite{kobayashi-etal-2022-diverse} \par
\textbullet~disagreement measure \cite{kobayashi-etal-2022-diverse} \par
\textbullet~correlation coefficient \cite{kobayashi-etal-2022-diverse} \par
\textbullet~R-QAE \cite{lee-etal-2020-generating} \par %
\textbullet~average length \cite{panagiaris-etal-2020-improving} \par
\textbullet~frequency of low-frequency words \cite{han-etal-2022-measuring} \par
\textbullet~demand-weighted utility \cite{khanuja-etal-2023-evaluating} \par %
\textbullet~F1 of a baseline coreference solver \cite{zhukova-etal-2022-towards} \par
\textbullet~Honoré’s R \cite{kosmajac-keselj-2019-twitter}  \par
\textbullet~Sichel’s S \cite{kosmajac-keselj-2019-twitter}  \par
\textbullet~Yule's K \cite{kosmajac-keselj-2019-twitter} 
\\ \hline
\end{tabular}
\caption{Families (in bold) of diversity measures in NLP non-conforming to \citeauthor{stirling_general_2007}'s framework. Relative quantification in the upper part. Number of papers per family in parentheses.}
\label{tab:non-conforming-measures}
\end{scriptsize}
\end{table}

\final{In this section, we discussed families of measures over our entire corpus.
If we break the corpus down based on the year of publication, as in Figure~\ref{fig:diversity-evolution-time}, we observe 
little variation, with the exception of variety, which saw a sharp increase in 2022 and 2023. We observed that this is
 related to the respective increase %
at Input and Data collection stages over the same time period
(Figure~\ref{fig:diversity-evolution-stage}).}

\begin{figure}
    \centering
    \resizebox{\linewidth}{!}{
    \input{images/diversity-evolution-time-alt.pgf}
    }
    \caption{Evolution in time of families of measures. Note that the data for 2024 ends on July 26th.
    \label{fig:diversity-evolution-time}}
\end{figure}

\section{Discussion}
\label{section:discussion}

Our examination of the interactions between \emph{why}, \emph{what}, \emph{where} and \emph{how} perspectives \final{allowed us to take a step back and gain a clearer understanding of diversity quantification in NLP. We found that normative assumptions behind diversity endorsement are usually implicit but can be excavated and divided into ethical and practical ones. Those are often intertwined and not always taken for granted. There are many different NLP areas where diversity matters and diversity measurement targets many different categories (mainly in-text and meta-linguistic ones).
Various stages of the ML pipeline are involved, but the Data collection, Input data and Output data stages dominate.  The apparent large number (over 100) of names of diversity measures, when comparatively examined, can be boiled down to about 80 measures (i.e. in some cases the same measure bears different names) and 13 families, 7 of them conforming to \citeauthor{stirling_general_2007}'s (\citeyear{stirling_general_2007}) framework. The two largest families -- dRichness and dAvePairwise -- when viewed from a certain level of abstraction, are actually unique measures. Overall, variety and disparity measures prevail, while balance measures remain relatively unexplored. There are also relatively few hybrid measures, and in particular no hybrids with disparity. Among disparity measures, many distance measures are used but few methods for aggregation of distances are examined; 84\% of disparity-related papers only use straightforward averaging. Non-conforming measures are roughly as many as conforming ones, the dTTR family dominates and relative quantification is popular.}

Our framework also revealed two main prototypical, highly distinct, scenarios. 
The first involves the area of %
\final{\emph{Resources and Evaluation}}. At
the Data collection pipeline stage, the selection of sources is guided by \emph{ethical} goals of inclusiveness and equality, and/or by \emph{practical} ones of ensuring performance. The diversity categories are \emph{meta-linguistic} and measures from the \emph{dRichness} family (variety) are used. Papers in this scenario range from as simple a quantification as three genres
by \citet{etienne-etal-2022-psycho} to as many as dozens of languages with two levels of categories by \citet{vylomova-etal-2020-sigmorphon}. The latter describes a shared task on morphological inflection in which diversity is operationalized in the choice of languages for the dataset: 90 languages from 34
genera, grouped into 15 language families. Families are used as proxies of different
morphological systems. The task is to generate inflected forms from a lemma and a set of features. The systems
have to
generalize across typologically distinct languages (also those unseen in training), many of which are low-resourced. %

The second scenario occurs in \emph{Generation}\final{, \emph{NLP Applications} and \emph{Machine Learning for NLP}} at the ``Output data'' stage, where \emph{in-text} categories are used for evaluation. Diversity is a \emph{practical} goal related to user expectations.
Often the \emph{one-to-many} scenario
occurs and the \emph{quality/diversity trade-off} has to be balanced. The measures are mostly from the \emph{dTTR} or \emph{dAvePairwise} 
families. For example, \citet{liu-etal-2023-pvgru} investigate response generation for multiturn dialogue in generative chatbots. Their objective is to enhance chatbot responses for diversity and relevance simultaneously. They use recurrent neural networks (whose basic transition structure is deterministic) into which they introduce a recurrent summarizing latent variable (to enable variability), to address the one-to-many challenge. The diversity of the generated answers is measured with \texttt{distinct-1/2} (\emph{dTTR}) and with ranking-based human evaluation (\emph{dHuman}).

Several comments can be made
on
the comparison between diversity quantification in NLP and in other fields (Sec.~\ref{sec:other-fields}). 

First, NLP shows a relatively weak awareness of the longstanding literature about diversity in other fields %
\citep[but see][]{kosmajac-keselj-2019-twitter,dunn-etal-2020-measuring,lion-bouton-etal-2022-evaluating}.  
\final{Conversely, in genomics, measures of clonal diversity mostly originate from ecology \cite{corre2023evaluation}. In economics, \citet{maignan2003-bio-ecological-diversity} systematically compare bio-ecological and socio-economic diversity measures, and \citet{davydov_simple_2016} relate the societal fractionalization index to its counterparts in ecology \citep{simpson_measurement_1949,hill_diversity_1973} and linguistics \citep{greenberg_measurement_1956}. 
In bibliometrics, standard diversity indicators are borrowed from ecology and economics \cite{mugabushaka_bibliometric_2016}.
Finally, \citeauthor{stirling_general_2007}'s (\citeyear{stirling_general_2007}) framework is applied (i) to quantify cultural and ethnic diversity in research evaluation \cite{gok_measuring_2026}, (ii) to characterize the degree of craft standardization in archaeology \citep{fragnoli_re-assessing_2021}, and (iii) 
 to quantify diversity in information networks with Hill numbers \citep{ramaciotti_morales_measuring_2021}.}

\final{Second, NLP shares other domains' tendency to often use richness (Sec.~\ref{sec:philo}) but there is a large difference in scale and in the way categories are derived. Many domains refer to a set of relatively few fixed  predefined categories:  (i) 5 to 12 electricity supply options \citep{stirling_diversity_1994} and 9 to 81 employment sectors \citep{ImbsWacziarg2003diversification} in economics, (ii) 12 religious and political orientations in sociology \citep{lieberson_measuring_1969}, (iii) 175 subject categories in publications' references in bibliometrics \citep{mugabushaka_bibliometric_2016,rafols_diversity_2010}, (iv) 4 rock types and up to 2200 species in mountain regions in geology \citep{rahbek_building_2019}, and (v) 10,000 clones (families of cells sharing the same unique insertion site) in gene therapy \citep{corre2023evaluation}.}
A relatively low number of pre-defined categories is also found in NLP, especially when meta-linguistic categories (languages, language families, genres) are concerned. But in-text categories (word types, \emph{n}-grams, syntactic subtrees, sentence clusters, etc.) emerge from the data rather than from a predefined classification. Their number is not fixed in advance and can be very large, e.g. over 200K unique n-grams \citep{park-etal-2023-dive}.

Third, NLP shows a relatively high number of cases where diversity measures are applied directly to elements, without apportionment into categories \citep[e.g., ][]{han-etal-2021-generating,gueuwou-etal-2023-jwsign,guo-etal-2024-curious}. This happens most often when utterances on input or output of the system are compared by similarity measures (\emph{dAvePairwise}, \emph{dVolume}, \emph{dDistOther}). Such scenarios can be seen as trivial cases of disparity measurement in which elements are identical to categories. Variety is then roughly
equivalent to dataset size (and not considered a diversity dimension) and balance is moot. 
\final{In economics and bibliometrics, by
contrast, elements are routinely apportioned into categories: units of
electricity output into supply options \citep{stirling_diversity_1994},
workers or value added into economic sectors \citep{ImbsWacziarg2003diversification}, and
references into subject categories
\citep{mugabushaka_bibliometric_2016, rafols_diversity_2010}.
}
Such cases reveal the tension between NLP, where continuous representations are prevalent, and other fields in which categorical modeling is standard. 

Let us finally note that looking only at ``diversity'' and ``diverse'' in titles bears the risk of missing important papers which use these terms in their bodies, or which use other related terms. To estimate this risk, we studied 100 extra papers containing \emph{diversity}, \emph{entropy}, \emph{variety}, \emph{balance}, and \emph{disparity} in paper bodies.%
\footnote{To select these 100 papers, we looked at papers in our time range containing at least one of these terms in their bodies and computed the average number of such occurrences. We then selected, for each term, 20 papers not yet in our corpus having more occurrences of this term than on average.}
We found that \emph{entropy} mostly occurred in \emph{cross-entropy} as loss function, not as a diversity measure. \emph{Diversity}, \emph{diverse}, \emph{disparity}, and \emph{variety} were either unrelated to diversity quantification or did not bring anything new to our framework. The case of \emph{balance} was more interesting. This term was sometimes used to characterize training or test corpora in which the distribution of classes was more or less balanced, which had consequences for the models' performance. The distribution was quantified but not aggregated into a single balance score. 
This aspect of balance has a long history in NLP but, interestingly, is not related to the concept of diversity in any of these papers.

\section{Recommendations for NLP}
\label{sec:recommandations}

The state of NLP with respect to diversity quantification appears in our survey as comparable to the first two stages of evolution in ecology: proliferation of measures and ``nonconcept'' (Sec.~\ref{sec:ecology}). A few efforts, like \citet{tevet-berant-2021-evaluating,lion-bouton-etal-2022-evaluating,yang-etal-2025-measuring,zhang-etal-2025-evaluating-evaluation,guo-et-al-benchmarking:tacl:2025} and ours, take first steps towards comparability of different measures. To pursue these efforts, we suggest that new papers addressing diversity quantification position themselves on the \emph{why}, \emph{what}, \emph{where} and \emph{how} perspectives, as defined in this survey. This includes answering the following questions: (i) Is diversity increase endorsed in your paper and if so, what are the motivating normative assumptions behind this stand? (ii) What are the categories and elements on which diversity is measured? (iii) At which stage of the processing pipeline do you quantify diversity? (iv) What is your selected diversity measure? What is its relation to other existing measures, and to variety, balance and disparity? (v) If balance is concerned, how to characterize the distribution of the elements into categories? (vi) If disparity is concerned, what is the underlying distance measure and the distance aggregation method? (vii) What are your reasons for choosing this particular measure and why is it a diversity measure (and not a performance measure for instance)? 
One of these perspectives may occasionally need extensions. For instance, recent papers propose LLM-as-a-judge instead of humans for scale-based diversity evaluation \cite{cao-etal-2025-perspective}, which extends the \emph{dOther} family.

While these steps towards a shared vocabulary are being taken, NLP should initiate in-depth debates about desirable properties of diversity measures. For instance, the status of measures non-conforming to Stirling's framework (Tab.~\ref{tab:non-conforming-measures}) should be questioned: maybe some of them, such as TTR, average length of an utterance or relative measures (\emph{dRelOverlap}, \emph{dRelDistrib} and \emph{dRelDist}) should not be considered measures of diversity but of related properties (e.g.\ difference or complexity). We should also consider the distance measures underlying disparities. Currently some of them, such as BLEU, are not formal distance metrics (e.g.\ they are non-symmetric). Also,  some methods of aggregating distances may have undesirable properties. In particular, computational tractability of disparities should be addressed.

One inspiration for these debates could come from Hill numbers, valued for their interpretability, as well as diversity profiles (Sec.~\ref{sec:ecology}). %
Ecological studies on disparities \cite{schleuter_users_2010,magneville_mfd_2022} could extend our understanding of distance aggregation methods, which still remain unexplored, as seen in Tab.~\ref{tab:conforming-measures}. On the other hand, we should address the particularities of language data, such as the prevalence of Zipfian distributions in the in-text diversity scenarios. On top of these efforts, diversity measures and their properties should be further studied and formalized.

Following \citet{sarkar_diversity_2010}, we also suggest that the choice of diversity measures be dictated by the normative assumptions behind research. This also calls for community debates. This could be approached by systematically integrating diversity as an evaluation dimension, e.g.\ in shared tasks and benchmarks.

\bibliographystyle{acl_natbib}
\bibliography{bibliography.bib,exp-012_diversity_diverse_2019-2024_ACL_2024-07-26.bib,bib.bib,disparity.bib,other.bib}

\appendix

\end{document}